\documentclass{article}
\usepackage[utf8]{inputenc}
\usepackage{graphicx}
\usepackage{float}
\usepackage{algorithm}
\usepackage{algpseudocode}
\usepackage{url}
\usepackage{subcaption}
\usepackage{amssymb}
\usepackage{authblk}
\usepackage{wrapfig}
\usepackage{todonotes}

\begin{document}

\title{An End-to-End Approach for \\Online Decision Mining and Decision Drift Analysis in Process-Aware Information Systems: Extended Version}

\author[1]{Beate Scheibel} 
\author[2]{Stefanie Rinderle-Ma} 
\affil[1]{Research Group Workflow Systems and Technology,
Faculty of Computer Science,
University of Vienna,
beate.scheibel@univie.ac.at}
\affil[2]{
Technical University of Munich, Germany; TUM School of Computation, Information and Technology
stefanie.rinderle-ma@tum.de
}
\date{}
\maketitle

\begin{abstract}

Decision mining enables the discovery of decision rules from event logs or streams, and 
constitutes an important part of in-depth analysis and optimisation of business processes.
So far, decision mining has been merely applied in an ex-post way resulting in a snapshot of decision rules for the given chunk of log data. Online decision mining, by contrast, enables continuous monitoring of decision rule evolution and decision drift. Hence this paper presents an end-to-end approach for the discovery as well as monitoring of decision points and the corresponding decision rules during runtime, bridging the gap between online control flow discovery and decision mining. 
The approach provides automatic decision support for process-aware information systems with efficient decision drift discovery and monitoring. For monitoring, not only the performance, in terms of accuracy, of decision rules is taken into account, but also the occurrence of data elements and changes in branching frequency. 
The paper provides two algorithms, which are evaluated on four synthetic and one real-life data set, showing feasibility and applicability of the approach. Overall, the approach fosters the understanding of decisions in business processes and hence contributes to an improved human-process interaction.
\end{abstract}

\section{Introduction}
\label{sec:intro}

Process mining and specifically decision mining allows for increased transparency of processes, which is crucial across all domains \cite{leewis_future_2020}. 
Decision mining is a part of process discovery, allowing for the discovery of decision points in a process model and the corresponding decision rules guarding that decision based on data elements \cite{sakr_decision_2018,rozinat_decision_2006}. 
A decision rule can consist of multiple conditions, which are usually of the form $v(ariable)\ op(erator)\ c(onstant)$, e.g., \textsl{temperature below 50°}. Conditions can be concatenated to form decision rules. Decision mining can be seen as a classification problem. Therefore the potential branches that can be chosen and executed are regarded as decision classes.

Existing decision mining methods \cite{sakr_decision_2018} are applied in an ex-post manner. However, especially when aiming at increased transparency, runtime analysis is particularly interesting, as information about decisions can be communicated to the user in almost real-time. In addition, runtime analysis allows for the prompt detection of decision drift, i.e., the manifestation of changing decision rules and decision points in event logs and streams.
Decision drift can occur due to errors or changes in the environment. Detecting drifts is important to ensure correctness and compliance of a process, i.e., assuring that the drift occurred intentionally and not due to errors. This is crucial across domains such as manufacturing to ensure quality of products and health care to ensure quality of patient care. 
This is especially relevant as 
\textsl{``[e]ffective decision making -- that is connected, contextual and continuous -- results in a host of business benefits, including greater transparency, accuracy, scalability and speed} \cite{Gartner2021}. 
Our previous work \cite{scheibel_online_2022} introduced an approach for detecting decision rule changes during runtime. However, the approach has several limitations. It is assumed that decision points are already known. Therefore the approach cannot be used as end-to-end approach, i.e. decision discovery methods have to be applied complicating analysis during runtime. 
In addition, the definition of decision drift has been limited to changes in decision conditions, neglecting decision point changes.

\begin{figure}[htb!]
    \centering
    \includegraphics[scale=0.3]{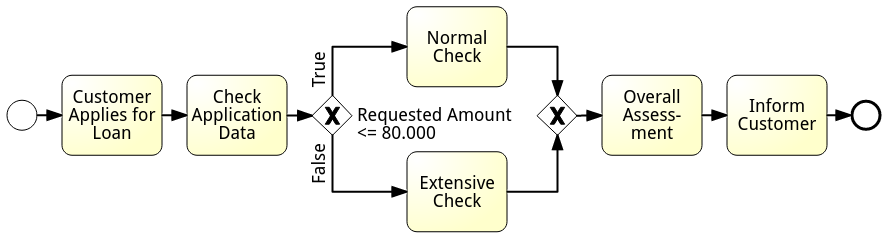}
    \caption{Running Example, Loan Application}
    \label{fig:running}
\end{figure}

As running example consider a simplified loan application process depicted in Fig. \ref{fig:running}. A customer applies for a loan, the application data is checked for completeness. Then either a normal or extensive check is performed. The results of the check contribute to the overall assessment that results in either rejection or acceptance of the loan. Lastly, the assessment is communicated to the customer. The exemplary process includes one decision point, i.e., whether a normal or extensive check is necessary. Multiple data elements are part of the process and serve as basis for the decision, e.g., the requested amount.

Standard decision mining techniques result in the following decision rule:
\noindent $\ IF\ amount\_loan <= 80.000\ THEN\ Normal\ ELSE\ Extensive$.
The decision depends on data element \textsl{amount\_loan}. Assume that during process execution changes can occur, e.g., a change in regulation leads to stricter checks and therefore an extensive check is already required for any amount greater than $50.000$.
Other changes might include additional data that becomes available during process execution and can be used to more accurately mine decision rules, e.g., \textsl{income}, the addition of an additional branch (class) at the existing decision point, e.g., a branch \textsl{Simple Check} is added, or the addition of a new decision point, e.g., a customer is handled differently depending on whether the assessment resulted in rejection or acceptance of the application.

A comprehensive, end-to-end decision mining and monitoring approach should be able to mine and monitor decision rules and decision drifts during runtime
as soon and as accurately as possible. End-to-end requires minimum involvement of users for setting up the algorithms and providing meaningful results. Online requires continuous mining with limited storage and appropriate handling of outdated data. These requirements can be addressed based on the following questions: RQ1: What exactly is decision drift and when/why does it happen? RQ2: How to mine decision points, rules, and drifts in a connected and continuous way without prior knowledge of the process model? RQ3: How to deal with limited storage? and RQ4: How to deal with outdated data which might become useful or even detrimental for mining current decision rules?

To address RQ1--4, this paper derives and discusses a definition of decision drift and its triggers based on literature ($\mapsto$ RQ1). In order to meet RQ2, the approach is designed to only require an event stream as input to provide users with information about current decision points, the corresponding rules, and potential drifts. The presented approach is comprehensive as it mines decision points, rules, and drifts in a connected way  ($\mapsto$ RQ2). 
For this, data values as well as the frequency of branching conditions are taken into consideration as indicators for decision drift. 
If drift is detected, users are notified about the drift and changed decisions, and can check if these changes are intentional ($\mapsto$ RQ2).
To account for limited storage and outdated data, a window based approach is taken, where only the recent data is taken into account ($\mapsto$ RQ3 and RQ4).
Overall, used in conjunction with a process aware information system (PAIS), that continually provides new event data in the form of an event stream, the approach provides continuous and increased transparency for users. 

Section \ref{sec:analysis} discusses and defines decision drift concepts. Section \ref{sec:approach} describes the algorithms, which are evaluated in Sect. \ref{sec:eval}. Related work is discussed in Sect. \ref{sec:rw} and a conclusion is provided in Sect. \ref{sec:conclusion}.

\section{Decision Drift: Definition and Analysis}
\label{sec:analysis}

Decision drift refers to different kinds of changes affecting decisions in a process. A decision is defined by a corresponding decision point in a process model (control flow) and the associated decision rule defining which branch is chosen based on process data (data flow). Decision drift, consequently, can occur due to control flow change, data flow change, and changes to the decision rule itself. Hence, in the following, we analyze state-of-the art approaches for process change (patterns), changes of data and decisions, as well as concept drift, aiming at achieving an understanding and definition of decision drift.

\cite{weber_change_2008} provide a framework of process change patterns referred to as adaptation patterns (AP). The following APs are relevant for decision drift: AP1 describes the insertion of process fragments, including a conditional insert, i.e., an activity is inserted into a process model together with a condition, i.e., decision point. AP2 refers to the deletion of a process fragment, which can also entail the deletion of a decision point. AP8 describes the addition of a loop, which also involves the addition of a decision point. AP10 refers to the addition of a decision point in the process. Lastly, AP13 refers to modifications of decision rules.

\cite{hasic_decision_2020} define change patterns in Decision Model and Notation (DMN) models. DMN models consist of a decision requirement diagram (DRD), depicting input of decisions and the dependencies between elements. Elements can either be decision nodes or input nodes. Each decision node can be represented by a table, including multiple decision rules. A decision rule consists of combinations of input and output variables, i.e., decision classes. \cite{hasic_decision_2020} propose four change categories. First, change within decision rules, i.e., the decision table changes by including or deleting input or output, or a change in the decision logic. Second, change on decision rules in their entirety, i.e., the inclusion or exclusion of a decision rule from a decision table. Third, change of the decision nodes in the DRD, i.e., deleting/adding a decision node (consisting of multiple decision rules). And lastly, change of the input data nodes in the DRD, i.e., including or excluding data as input.

Summarizing the literature analysis results, the following decision changes can potentially occur in a process: changes of data values in a condition, addition/deletion of a condition a decision rule, addition/deletion of data elements in a decision rule, addition/deletion of decision classes in a decision rule, and addition/deletion of decision points in a process model. 

Concept drift \cite{king_handling_2011} describes changes in processes with regards to their manifestations in the process execution (logs), i.e., sudden, recurring, gradual, and incremental drift. Accordingly, decision drift can be understood as manifestation of decision changes in process execution logs (ex post) or event streams (online). 
With respect to their detection and monitoring, 
decision drifts can be further classified in changes of decision rules (incorporating changes in conditions), decision classes, and decision points. Changes on higher levels, i.e., decision points, classes or rules, can also entail changes on lower levels, i.e., conditions, rules or classes. In \cite{scheibel_online_2022}, a significant drop in accuracy when predicting newly incoming instances
is used as sign that a decision drift occurred. \cite{lu_detecting_2021} suggests detecting changes in decision rules by monitoring the branching frequency, i.e., how often a specific branch is chosen. 
\cite{cappiello_detecting_2019} looks at changes in data values to determine if a concept drift occurred.
Overall, decision drift can be detected based on (1) decreased performance, (2) changing branching frequency and (3) changes in data elements or data ranges.
A comprehensive decision drift analysis approach should be able to monitor occurrence of (1)--(3) in order to detect decision drift.

\section{End-to-End Runtime Decision Mining, Monitoring, and Decision Drift Detection Approach}
\label{sec:approach}
\begin{figure}
    \centering
    \includegraphics[scale=0.5]{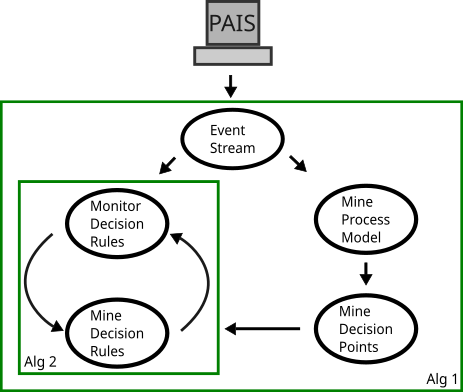}
    \caption{End-to-End Approach -- Overview}
    \label{fig:overview}
\end{figure}

The overall approach is depicted in Fig. \ref{fig:overview}. The input is an event stream, emitted by, e.g., a PAIS, which is used to mine a process model using online process discovery methods (for an overview of online methods see \cite{van_der_aalst_streaming_2022}). The process models are the basis for determining the decision points and the corresponding decision classes, which, in turn, are the basis to mine decision rules for each of the decision points. The decision rules are continuously monitored, using newly incoming events. If either the performance, the frequency of taken branches, or the data value ranges change, we assume that decision drift occurred and remining is performed. Therefore the approach consists of two continuous processes: first, the process discovery part continues considering newly occurring events and remining the process model if necessary. If this leads to new or changed decision points, decision rules are remined as well (cf. Alg. \ref{alg:dmma}). Second, the existing decision rules are continually checked for compliance with newly incoming events and remined if necessary (cf. Alg. \ref{alg:monitoring}). 

Algorithm \ref{alg:dmma} reflects the overall framework, combining existing methods for process discovery and decision point analysis with new techniques for decision rule mining to achieve a fully automatic end-to-end approach. 
Algorithm \ref{alg:dmma} calls Alg. \ref{alg:monitoring} for decision rule and remining in case of decision drifts.

\begin{algorithm}[htb!]
 	\caption{Runtime Decision Mining, End-to-End Approach}
 	\label{alg:dmma}
	\hspace*{\algorithmicindent} \textbf{Input: Queue of Elements From Event Stream, Grace Period} \\
    \hspace*{\algorithmicindent} \textbf{Output: Decision Points, Decision Rules for Each Decision Point}
 	\begin{algorithmic}[1]
 	\State Trace\_Dict = \{\}, Window Size WS = Grace Period
 	\While{Element in Queue}
 	\State Adapt Directly-Follows-Graph(Element), using Lossy Counting  	\State HN = Make Heuristics Net(Directly-Follows-Graph)
 	\If{New HN} \Comment{Find Decision Points}
 	\State Find Places with Multiple Outgoing Arcs as PS
    \State Discover Respective Classes (i.e. next activities) 
    \State DPS\_Data =  Dictionary with Decision Point as Key and Empty Values 
 	\EndIf
 	\State Get Data from Element and Store in Trace\_Dict
 	\If{Current Event in DPS\_Data.Keys}
 	\State Store Data Up to This Point From Trace\_Dict in DPS\_Data[Current Event]
 	\If{DPS\_Data[Current Event] $>$ WS}
 	\State Remove Oldest Instance
 	\EndIf
 	\If{Current Event is the Last Decision Point of Instance}
 	\State Remove Instance From Trace\_Dict
 	\EndIf
 	\EndIf
 	\If{Grace Period Finished} \Comment{Initial Rule Mining}
 	\For{DP in DPS\_Data.Keys}
    \State Data = From DPS\_Data[DP] 
    \State Build Decision Tree
    \State Build ADWIN Models for Data, Decision Classes, Accuracy
    \State Store Decision Tree and ADWIN Models in DMS
    \EndFor
 	\EndIf
 	\If{Current Event in DPS\_Data.Keys AND Initial Mining Finished}
 	\State DMS, WS = Monitoring(Current Event, DPS\_Data, DMS) see Alg. \ref{alg:monitoring}  
 	\EndIf
 	\EndWhile
 	\end{algorithmic} 
 \end{algorithm} 
Algorithm \ref{alg:dmma} works as follows: as soon as a new event occurs it is stored in a queue, which continues to store newly incoming events, while the next phases of the approach are executed in parallel. Therefore, even while process models and decision rules are discovered, new incoming elements are not lost. Each event is stored in the queue and will be processed. In addition to the event stream, the initial grace period is needed as input, either set manually, or by default, a grace period of $200$ instances will be used. The appropriate grace period is oriented towards how frequently new instances arrive and how complex the contained data and underlying decision rules are expected to become. However, when testing the approach, we did not observe a significant impact by different settings of the grace period. 

Each new event is instantly stored as part of the directly-follows-graph, which contains two events and the count of how often this combination occurred. \textit{Lossy counting} and the \textit{S-BAR method} \cite{zelst_event_2018} are used, which continually drop less frequent combinations, thereby accounting for finite storage and concept drift.
The next step, the \textbf{process model discovery}, is realized using the Heuristics Miner (HM) \cite{DBLP:conf/cidm/WeijtersR11}. The HM is used as discovery technique, as only a directly-follows-graph is needed as input, whereas other algorithms often need defined start and end events. This is not trivial when dealing with runtime discovery, as it is not know when an instance is finished.

As output, a petri net is generated, which is used as input for the \textbf{decision point discovery}. Using the algorithm proposed in \cite{rozinat_decision_2006}, decision points, i.e., the places with multiple outgoing arcs, as well as the decision classes, i.e., the next occurring events, are discovered. A discovered decision point could look like this: $"Check application data": ["Normal Check", "Extensive Check"]$, i.e. the event before the decision as well as the decision classes are specified.

Up to this point, the only used data structure is the directly-follows-graph which is stored in a hash table, more specifically a dictionary. As soon, as decision points are mined, two additional hash tables are created. First, \texttt{Trace\_Dict} is a dictionary, where each new instance is stored, using the instance identifier as key and all events and corresponding data elements of the respective instance as values. Second, \texttt{DPS\_Data} uses decision points as key and every time an event occurs that was beforehand identified as a decision point, all available information in \textit{Trace\_Dict} up to this point for this instance, is stored as the value.

After the grace period finishes and decision points have been found, \textbf{decision rule mining} is performed, using the data stored in \texttt{DPS\_Data} for the decision points. Any decision mining method can be applied. Here, a CART decision tree is used. The mined rules for each decision point are stored in a dictionary \texttt{DMS} using the decision point as key and the current decision tree models as well as some statistics as values, for example the accuracy, that will be added in the monitoring phase. In addition, a new ADWIN instance is generated for the average accuracy, each decision class, and each data element that is part of the associated decision point and stored in \texttt{DMS}. ADWIN is a well-known approach for concept drift detection \cite{bifet_learning_2007}, where the window size is adapted according to the change of data in the window. ADWIN is used in the monitoring phase to detect changes.

For each newly occurring event that reflects a decision point, Alg. \ref{alg:monitoring} for \textbf{Monitoring and Remining} is called. 

With regards to limited storage, the directly-follows-graph in combination with the S-Bar method, is inherently storage efficient. As for the \texttt{Trace\_Dict}, instances are removed as soon as all decision points have occurred and therefore all data for this instance is already stored in \texttt{DPS\_Data}. As for \texttt{DPS\_data}, for each decision point, the most recent instances are kept, the exact number depends on the window size \textit{WS}. The window size is by default set as the grace period and therefore the same for all decision points. As soon, as a drift is detected at a decision point, the window size is set to the ADWIN window size for that decision point. See Alg. \ref{alg:monitoring} for details.

\label{sec:alg2}
\begin{algorithm}
 	\caption{Monitoring and Remining}
 	\label{alg:monitoring}
	\hspace*{\algorithmicindent} \textbf{Input: Decision Point DP, DPS\_Data, DMS} \\
    \hspace*{\algorithmicindent} \textbf{Output: DMS with Updated Decision Rules}
 	\begin{algorithmic}[1]
 	\State Compare Current Data with DPS\_Data
    \If{New Data Element}
    \State DMS[DP][Drift] = True
    \EndIf
 	\State Predict Class for Current Event using DMS
 	\State Calculate Overall Average of Accuracy, Data, Frequency
 	\State Add Averages to Respective ADWIN Models, Calculate Drift
 	\If{ADWIN Drift Detected}
 	\State DMS[DP][Drift] = True
 	\State WS = ADWIN.window
 	\EndIf
 	\If{DMS[DP][Drift] == True}
 	\State Remine Decision Model, Store Decision Model in DMS
    \State Reset Stored Data (Averages, ADWIN Models,...)
    \State DMS[DP][Drift] = False
 	\EndIf
 	\State Return DMS, WS
 	\end{algorithmic} 
 \end{algorithm} 

Algorithm \ref{alg:monitoring} builds on our previous work presented in \cite{scheibel_online_2022} and has been significantly extended and adapted. Instead of relying on changes in the performance of decision rules, the monitoring also includes, data elements, i.e. new data elements occurred at a decision point or ranges of data values changed and changes in the branching frequency. Branching frequency refers to the frequency that decision classes, i.e. the respective branch, is chosen. 
The function is called with an event that has been identified as a decision point, e.g., for the running example, a new event \textsl{Normal Check} following an event \textsl{Check Application Data} occurred, which is part of a decision point.
First, all data that occurred up to this point for this instance, and is stored in \texttt{Trace\_Dict}, is gathered. The names of the data elements are compared to the data elements that have occurred before at this decision point and have been stored in \texttt{DPS\_Data}.
E.g., up until the event \textsl{Normal Check} the data elements \textsl{requested amount} and \textsl{age} were logged for this instance. In \texttt{DPS\_Data} for the current decision point also only these two data elements are stored. Therefore no new elements have been detected.
If unseen data elements, e.g., a data element \textsl{income}, are discovered, the variable \texttt{DMS[DP][Drift]} is set to True and stored in \texttt{DMS}. 
Otherwise, the class for the current decision point is predicted and compared to the actual class, to calculate the current accuracy, which is used in the next step to calculate the average decision rule accuracy.

The drift detection method ADWIN \cite{bifet_learning_2007} is employed in order to detect whether a drift has occurred, either in the performance, i.e., the accuracy of the decision rules, the data values, or the branching frequency. ADWIN is the basis for window-based concept-drift detection methods such as \cite{hassani_concept_2019,maaradji_fast_2015} and compares statistics between windows to check if these are significantly different, i.e., a drift happened. As setting the window sizes manually is not trivial, the sizes are chosen according to the amount of changes in the data. If the data is stationary, the window is increased to improve accuracy. If drift occurs, the window is decreased. \cite{lu_detecting_2021} propose a method for identifying changes in process models based on changes in the branching frequency ex-post. This change detection method is not able to work with event streams. Therefore, we opted for the ADWIN approach that is specifically optimized for runtime analysis.

For the end-to-end approach presented in this paper, the average decision rule accuracy is calculated each time the monitoring function is called, using the overall accuracy and the overall number the function was called, which are both stored in \texttt{DMS}. The same is done to calculate the average branching frequency, i.e., the average percentage which branches are taken, i.e. which classes are chosen. For the running example, \textsl{Extensive Check} is on average performed for $30\%$ and \textsl{Normal Check} for $70\%$ of instances, but then the averages change to $40\%$ and $60\%$ respectively, which could be a sign that decision drift occurred. In addition, the average value for all data elements are calculated. Here, the average \textsl{requested amount} could be $45.000$, whereas the average age is $39$. 

The calculated averages are used as input to the ADWIN models.
In Alg. \ref{alg:dmma}, ADWIN models for the accuracy, the decision classes and data values are built at each decision point and stored in \texttt{DMS}. The calculated averages are added to the respective models, which then calculate whether a drift occurred.
If a drift is detected, \texttt{DMS[DP][Drift]} is set to True. The ADWIN window size from the model where drift was detected is set as window size. The window size is used to control the maximum size of \texttt{DPS\_Data}, i.e., if the ADWIN window size is, for example, $500$ after a drift was detected, no more than $500$ instances are stored for that decision point in \texttt{DPS\_Data}. The First In - First Out principle is applied, i.e., the oldest instance is removed as soon as the maximum size is reached. This allows to dynamically increase and decrease the amount of instances stored, which is necessary as storage is limited and outdated data should not be used for remining decision rules. As the window decreases when drift is detected, only the more recent instances are used to remine.

If \texttt{DMS[DP][Drift]} is set to True, a new decision model is mined and stored in \texttt{DMS}. Lastly, all corresponding data, e.g., the calculated averages and ADWIN models are reset.

\section{Evaluation}
\label{sec:eval}

The approach was implemented using python and as available online \footnote{\url{https://github.com/bscheibel/dmma-e}}.

As we propose, to the best of our knowledge, the first end-to-end runtime decision mining approach, the evaluation does not contain a comparison to other approaches. Instead, the general feasibility and applicability of the approach are evaluated. The requirements for data sets to be used for evaluation are:

\begin{enumerate}
    \item Process-based data set, e.g., an event log or stream
    \item Underlying process model contains one or more decisions
    \item Decisions are based on numeric data attributes
    \item Decision drifts occur
    \item Available ground truth: decision rules and drifts are known
\end{enumerate}

Requirements 1-3 are necessary to be able to use the approach at hand. Requirement 4 is prerequisite to show the ability of the approach to detect different kinds of drift. Requirement 5 enables the validation of the results, i.e., to check whether the detected decision rules and drifts correspond to reality. We analyzed publicly available real-life data sets from the BPI Challenge\footnote{\url{https://www.tf-pm.org/competitions-awards/bpi-challenge}} and from\footnote{\url{https://data.4tu.nl/}} along Requirements 1--5. The BPIC17 log fulfills Requirements 1--3 and the data elements are named to allow intuitive interpretation. Hence, the BPIC17 log is chosen for evaluating the applicability of the approach. In order to show its feasibility, we start with 
four synthetic data sets (SD) for which we know the ground truth.  The evaluation results contain the average accuracy and for the synthetic data sets SD I--IV the number of instances from a decision drift until new decision rules are remined.

\subsection{Feasability: Synthetic Datasets}
SD I--IV are based on the running example depicted in Fig. \ref{fig:running} and contain the following decision drifts: I) value changes in a condition, II) additional data elements in a decision rule, III) additional branch for a decision point and IV) an additional decision point. 
The corresponding datasets have been created using random variations. Each data set consists of $5000$ instances and is stored in a CSV file. In a pre-processing step, the contents are stored in a queue (event by event) in order to simulate an event stream. 
SD I--IV together with the complete evaluation results, including all decision rules, are available online\footnote{\url{https://github.com/bscheibel/dmma-e}}.

\begin{figure}[ht] 
  \begin{subfigure}[b]{0.5\linewidth}
    \centering
    \includegraphics[width=0.85\linewidth]{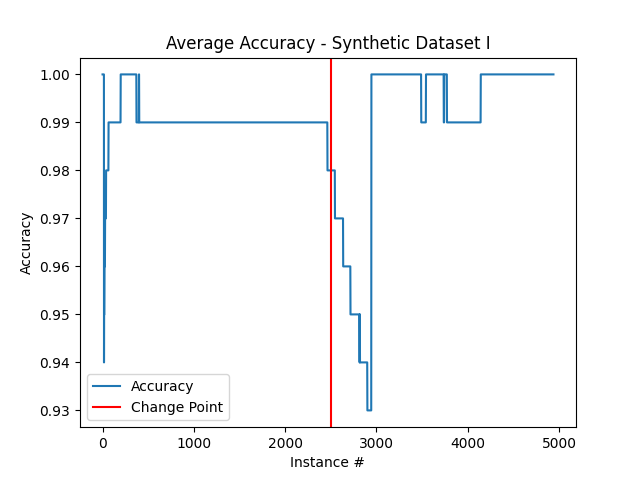} 
    \caption{SD I: Decision Rule Change.} 
    \label{fig7:a} 
    \vspace{4ex}
  \end{subfigure}
  \begin{subfigure}[b]{0.5\linewidth}
    \centering
    \includegraphics[width=0.85\linewidth]{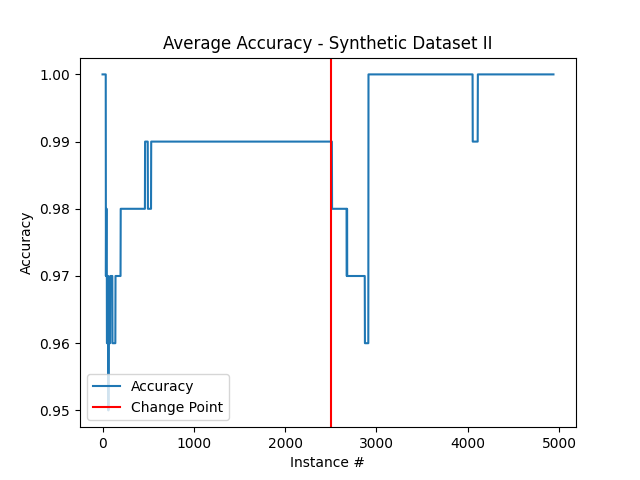} 
    \caption{SD II: Decision Rule Change.} 
    \label{fig7:b} 
    \vspace{4ex}
  \end{subfigure} 
  \begin{subfigure}[b]{0.5\linewidth}
    \centering
    \includegraphics[width=0.85\linewidth]{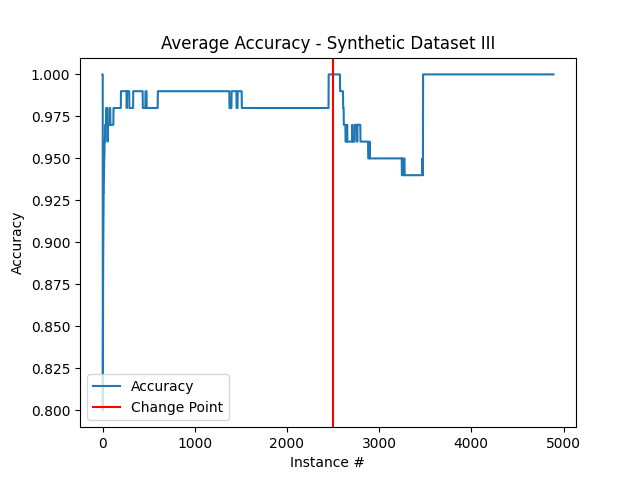} 
    \caption{SD III: Decision Class Change.} 
    \label{fig7:c} 
  \end{subfigure}
  \begin{subfigure}[b]{0.5\linewidth}
    \centering
    \includegraphics[width=0.85\linewidth]{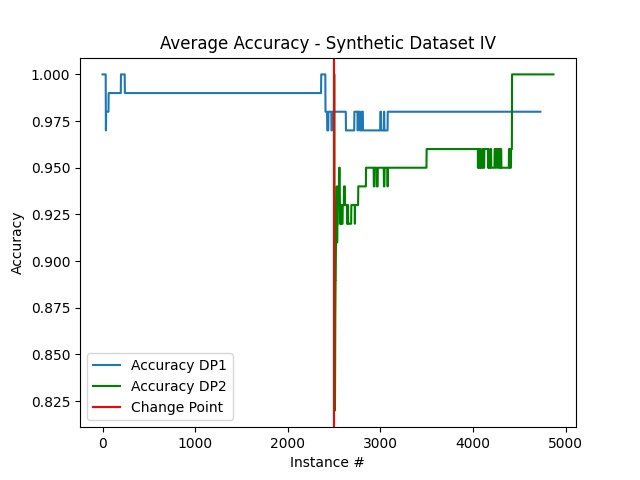} 
    \caption{SD IV: Decision Point Change.} 
    \label{fig7:d} 
  \end{subfigure} 
  \caption{Evaluation of SD I-IV.}
  \label{fig:eval_sd} 
\end{figure}

\noindent\textbf{SD I} reflects a decision rule change.
The initial rule: \\
\noindent $IF\ amount\_loan <= 80.000\ THEN\ Normal\ ELSE\ Extensive$ \\
is changed to: \\
\noindent $IF\ amount\_loan <= 50.000\ THEN\ Normal\ ELSE\ Extensive$ \\
at instance $2500$.
The overall average accuracy is $0.99$. The drift was detected $505$ instances after it occurred, at instance number $3005$, and the decision rule was remined. The result can be seen in Fig.  \ref{fig:eval_sd}a. The accuracy continually decreases after the drift occurred, immediately after remining at instance number $3005$, a sharp increase in accuracy can be seen.

\noindent\textbf{SD II} simulates an additional data element that is added to the decision rule. The initial decision rule is:\\
\noindent $IF\ amount\_loan <= 80.000\ THEN\ Normal\ ELSE\ Extensive$ \\
The data element \textsl{income} is added to the decision rule at instance number $2500$ and the rule is therefore changed to: \\
\noindent $IF\ amount\_loan <= 80.000\ AND\ income\ > 3000\ THEN\ Normal\ ELSE\ Extensive$ \\
To overall average accuracy is $0.99$. The drift was detected $473$ instances after it occurred, at instance number $2973$, and the decision rule was remined. The result can be seen in Fig. \ref{fig:eval_sd}b. The accuracy decreases after the drift occurred. After remining the accuracy increases.

\noindent\textbf{SD III} includes a decision class change, i.e., at first, only two decision classes are part of the decision point:

\noindent $IF\ amount\_loan\ <= 70.000\ THEN\ Normal \\
IF\ amount\_loan > 70.000\ THEN\ Extensive$ \\
then the additional class \textsl{Simple Check} is added at instance number $2500$: \\
\noindent $IF\ amount\_loan <= 30.000\ THEN\ Simple \\
IF\ amount\_loan > 30.000\ AND\ amount\_loan <= 70.000\ THEN\ Normal \\
IF\ amount\_loan > 70.000\ THEN\ Extensive$ 

The average accuracy is $0.98$. The drift was detected $62$ instances after it occurred, at instance number $2562$, and the decision rule was remined. However, the mined rules did not reflect the new rule accurately, therefore a second remining occurred at instance number $3585$. The result can be seen in Fig. \ref{fig:eval_sd}c. The accuracy decreases after the drift occurred. After the first remining the accuracy still decreases, whereas the second remining leads to an increase.

\noindent\textbf{SD IV} 
represents a decision drift in the form of an additional decision point. After \textsl{Overall Assessment}, two alternative branches are inserted: \textsl{Write Acceptance Letter} and \textsl{Write Rejection Letter}, before the branches are joined and the event \textsl{Inform Customer} occurs. 
\\
\noindent DP1: $IF\ amount\_loan <= 80.000\ THEN\ Normal\ ELSE\ Extensive$ \\
After instance number $2500$, a second decision point is added:\\
\noindent DP2:
$IF\ risk\_level <\ 4\ AND\ amount\_loan < 80.000\ THEN\ Write\ Acceptance\ Letter\\
ELSE\ IF\ risk\_level <=\ 1\ AND\ amount\_loan => 80.000\ THEN\ Write\ Acceptance\ Letter\\
ELSE\ Write\ Rejection\ Letter$

The overall average accuracy for DP1 is $0.98$ and for DP2 $0.96$. The drift was detected $133$ instances after it occurred, at instance number $2622$ and the decision rules were remined, including the new decision point. The result can be seen in Fig. \ref{fig:eval_sd}d. The accuracy for the first decision point remains relatively constant. However, after remining, the accuracy for the second decision point is added in the plot. The accuracy for the second decision point is relatively low at first. However, at instance number $4552$
another remining for the second decision point occurs, resulting in increased accuracy. 

\subsection{Applicability: BPIC17}

\begin{figure}[htb!]
    \centering
    \includegraphics[scale=0.2]{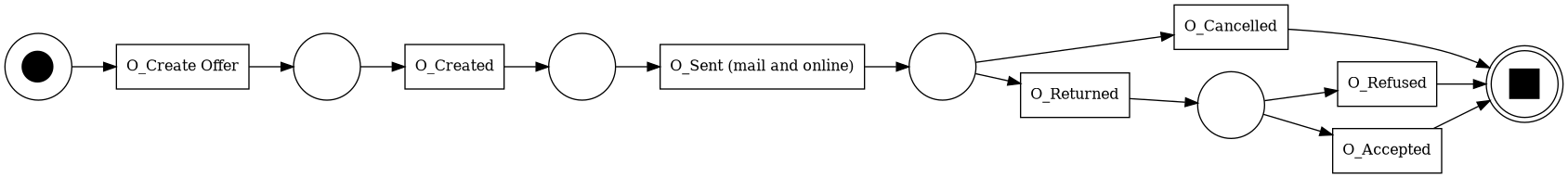}
    \caption{BPIC17: Mined Process Model.}
    \label{fig:bpic17}
\end{figure}

\begin{wrapfigure}[13]{r}{0.5\textwidth}
    \vspace{-5mm}
    \centering
    \includegraphics[width=0.5\textwidth]{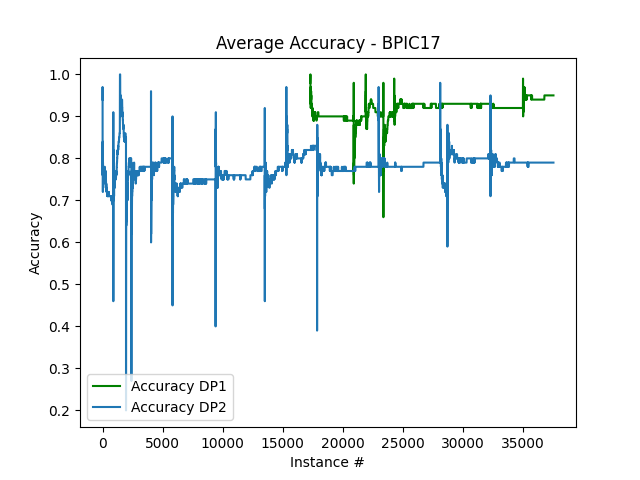}
    \caption{BPIC17 Results.}
    \label{fig:bpic17_eval}
\end{wrapfigure}

The BPIC17 data set\footnote{\url{https://data.4tu.nl/articles/dataset/BPI_Challenge_2017/12696884}} consists of a loan application process from a financial institute, including loan applications and offers, where each application can contain multiple offers. For the evaluation, the offer data set was used. Before applying the approach, pre-processing was done to simulate an event stream.

Fig. \ref{fig:bpic17} shows the final process model with two decision points. The process model was remined throughout the process: in the beginning it contained only one decision point. Figure \ref{fig:bpic17_eval} shows the accuracies of the decision points. The average accuracy for the first decision point is $0.78$ and for the second decision point $0.91$. In total, remining occurred $19$ times, $14$ times for the first decision point and $5$ times for the second decision point. 
Both decision rules include the data elements \textsl{NumberOfTerms}, i.e. the number of payback terms agreed to, and \textsl{CreditScore}. For example, the last mined decision rule for decision point 2, i.e., if the offer was refused or accepted, contains the condition that the offer is accepted if the CreditScore is above $324$. The exact values in the decisions change with each remining.

\subsection{Discussion}
\label{sec:discussion}

The evaluation of the synthetic datasets shows that the approach is feasible and able to discover different kinds of decision drift during runtime. The full results also show that the mined decision rules are equal to the underlying rules. For the real-life dataset, the evaluation shows that the approach is able to work with real-life data. However, the BPIC17 dataset probably did not encompass any decision drifts, and the frequent remining was executed rather due to insufficient data than to mine accurate decision rules. 
Additional evaluations in real-life settings will be part of future work to show the generalisability of the approach. 
In terms of interpretability, the output consists of textual decision rules, which enables manual interpretation and analysis. The approach can also be easily adapted to work with different kinds of decision mining approach, see Sect. \ref{sec:rw}, enabling the inclusion of e.g. categorical or time-series data as input.

\noindent\textbf{Limitations and threats to validity:}
The approach might not work for cases where drifts are happening very frequently. If lots of unfinished traces occur, this could lead to storage build-up. In addition, only sudden drifts have been tested in the evaluation. Theoretically the approach should also work for incremental and gradual drifts, however this would probably include frequent remining until a stable decision rule is discovered. Furthermore, the current definition of data change is a change in the average, this is of course a very restricted definition.

\section{Related Work}
\label{sec:rw}

The first decision mining approach \cite{rozinat_decision_2006} includes an algorithm for detecting decision points and the corresponding decision classes from a Petri Net as well as classification techniques to mine the  decision rules. Subsequent approaches focus on specific aspects of decision mining, e.g., including overlapping rules \cite{mannhardt_decision_2016}, incorporating decision rules based on linear relationships between variables \cite{de_leoni_discovering_2013}, 
or mining decision rules based on time series data \cite{scheibel_decision_2022} (for an overview see \cite{sakr_decision_2018}). Existing approaches employ ex-post algorithms. Recently, online or runtime analysis is gaining traction for online process discovery \cite{burattin_online_2015,navarin_towards_2020}, conformance checking \cite{koenig_compliance_2019}, drift detection \cite{hutchison_online_2012,stertz_process_2018}, and predictive process monitoring \cite{pauwels_incremental_2021}. Especially, drift detection, partly overlaps with decision drift analysis as changes in decision points are part of control flow drift. 
For process discovery, \cite{navarin_towards_2020} introduce an approach to mine data-aware declarative process models, i.e., constraints, from event streams. The approach is similar as constraints could also be seen as decision rules, but still quite different as constraints do not translate straightforwardly to decision points and rules. \cite{park_explainable_2022} present an approach for predictive decision mining for operational support. None of these approaches include decision drift analysis, the remining of decision points and rules, and the textual generation of decision rules. \cite{lu_detecting_2021} propose a method for identifying decision rule changes based on changes in the branching frequency. This is done ex-post and neither decision point discovery nor remining are part of the approach. However, part of the approach is included in this approach for detecting change. 
Our previous work \cite{scheibel_online_2022} assumes that decision points are already known and no changes of decision points occur. Hence, this work constitutes a significant extension of \cite{scheibel_online_2022}.

\section{Conclusion}
\label{sec:conclusion}

This paper presents an end-to-end approach for mining decision rules during runtime, as well as monitoring of decision drift, and updating decision points and the associated decision rules if necessary. Decision drift encompasses different changes with regards to decisions in processes, i.e., changes in decision rules, decision classes, and decision points. The change detection is based on the drift detection method ADWIN and monitors the performance, the branching frequency as well as data values for changes. The approach is optimized for runtime use, i.e., limited storage as well as forgetting outdated data is taken into account. An event stream generated by, e.g., a PAIS is used as input. 
The output comprises textual decision rules for each discovered decision point, that are updated as soon as decision drift is detected, as support for users to evaluate if these changes are intentional. This enables increased transparency and can be used as basis for process enhancement. The evaluation shows that the approach is able to detect different kinds of decision drift with high accuracy and to work with real-life data. However, further testing is planned for future work as well as the analysis of drift patterns, root-cause analysis, and drift prediction.\\\\

\noindent\textbf{Acknowledgements}

\noindent This work has been partly supported and funded by the Austrian Research Promotion Agency (FFG) via the Austrian Competence Center for Digital Production (CDP) under the contract number 881843 and the Deutsche Forschungsgemeinschaft (DFG, German Research Foundation) -- project number 277991500.

\end{document}